\documentclass[runningheads]{llncs}
\usepackage[T1]{fontenc}
\usepackage{graphicx}

\usepackage{amsmath}
\usepackage{bm}
\usepackage{esvect}
\usepackage[table]{xcolor} 
\usepackage{colortbl} 
\usepackage{amssymb}
\usepackage{float}

\usepackage{booktabs}
\usepackage{multirow}
\usepackage{algorithm}
\usepackage{algpseudocode}
\usepackage{amsmath}
\usepackage{cleveref}
\usepackage{graphicx,verbatim}
\begin{document}
\title{MicroMIL: Graph-Based Multiple Instance Learning for Context-Aware Diagnosis with Microscopic Images}
%
%
\author{
JongWoo Kim\inst{1}\textsuperscript{*}\index{Kim, JongWoo} \and
Bryan Wong\inst{1}\textsuperscript{*}\index{Wong, Bryan} \and
Huazhu Fu\inst{2}\index{Fu, Huazhu} \and
Willmer Rafell Quiñones Robles\inst{1}\index{Quiñones Robles, Willmer Rafell} \and
Young Sin Ko\inst{3}\index{Ko, Young Sin} \and
Mun Yong Yi\inst{1}\textsuperscript{\dag}\index{Yi, Mun Yong}
}
\authorrunning{J. Kim and B.Wong et al.}
\titlerunning{MicroMIL}
%

\institute{
Korea Advanced Institute of Science and Technology, Daejeon, South Korea\\
\texttt{\{gsds4885, bryan.wong, munyi\}@kaist.ac.kr}
\and
Institute of High Performance Computing, Agency for Science, Technology and Research (A*STAR), Singapore\\
\and
Seegene Medical Foundation, Pathology Center, Seoul, South Korea\\
}

\maketitle              
%
\begingroup
\renewcommand{\thefootnote}{*}
\footnotetext{Co-first authors with equal contribution.}
\renewcommand{\thefootnote}{†}
\footnotetext{Corresponding author.}
\endgroup

\begin{abstract}
Cancer diagnosis has greatly benefited from the integration of whole-slide images (WSIs) with multiple instance learning (MIL), enabling high-resolution analysis of tissue morphology. Graph-based MIL (GNN-MIL) approaches have emerged as powerful solutions for capturing contextual information in WSIs, thereby improving diagnostic accuracy. However, WSIs require significant computational and infrastructural resources, limiting accessibility in resource-constrained settings. Conventional light microscopes offer a cost-effective alternative, but applying GNN-MIL to such data is challenging due to extensive redundant images and missing spatial coordinates, which hinder contextual learning. To address these issues, we introduce \textbf{MicroMIL}, the first weakly-supervised MIL framework specifically designed for images acquired from conventional light microscopes. MicroMIL leverages a representative image extractor (RIE) that employs deep cluster embedding (DCE) and hard Gumbel-Softmax to dynamically reduce redundancy and select representative images. These images serve as graph nodes, with edges computed via cosine similarity, eliminating the need for spatial coordinates while preserving contextual information. Extensive experiments on a real-world colon cancer dataset and the BreakHis dataset demonstrate that MicroMIL achieves state-of-the-art performance, improving both diagnostic accuracy and robustness to redundancy. The code is available at \url{https://github.com/kimjongwoo-cell/MicroMIL}

\end{abstract}
\section{Introduction}
Cancer remains a leading global cause of mortality, necessitating advancements in diagnostic technologies to enhance early detection and improve survival rates. Whole-slide imaging (WSI) has emerged as a transformative tool in digital pathology, offering high-resolution insights into tissue morphology and disease-related anomalies \cite{Evans2015,Iyengar2021}. However, WSIs require substantial infrastructure due to their high acquisition costs, memory demands, and lengthy processing times, making them less practical in resource-limited settings \cite{Spanhol2016,Alkassar2021}.

\begin{figure}[H]
\centering
\includegraphics[width=0.9\linewidth, trim=0 180 0 0, clip]{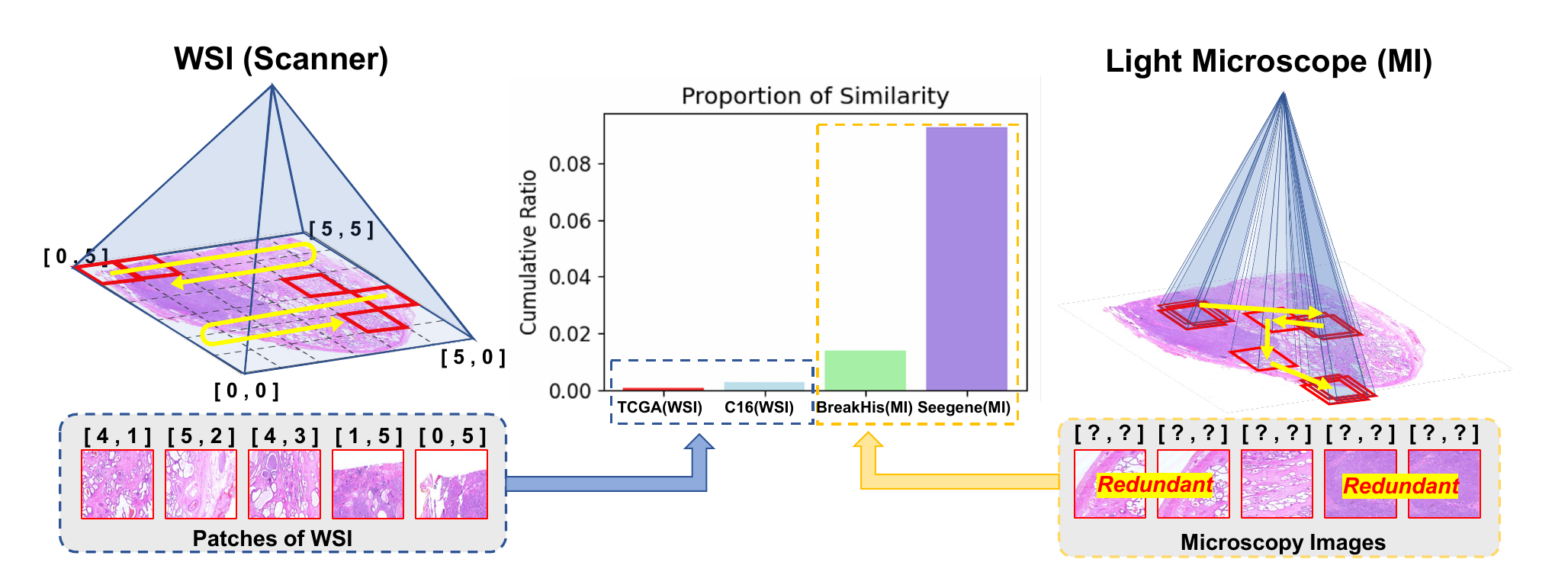}
\caption{\textbf{Left:} Valid patches from WSI (scanner) are acquired using a sliding-window approach and have absolute positions. \textbf{Right:} Light microscopy images lack known positions and contain many redundancies due to subjective capture by pathologists. \textbf{Middle:} Ratio of image pairs exceeding the redundancy threshold (0.995).}
\label{figure1}
\end{figure}

While WSI scanners are becoming more accessible, light microscopes remain far more widely used, especially in low-resource settings. Microscopy images thus offer a practical and cost-effective alternative for enabling AI-driven diagnostic solutions in diverse healthcare contexts \cite{sangameswaran2022maiscope}. Low-cost optical microscopes tailored for low- and middle-income countries continue to be developed and microscopy-based diagnostics remain the clinical standard \cite{Motivation_Microscope_1,Motivation_Microscope_2}. 

Despite these advantages, as illustrated in Figure \ref{figure1}, light microscopy images pose unique challenges, including the \textbf{absence of absolute spatial coordinates} due to manual acquisition by pathologists and \textbf{significant redundancy} caused by multiple image captures. To highlight these issues, we compare microscopic and WSI datasets (TCGA NSCLC\footnote{https://www.cancer.gov/tcga} and Camelyon16 \cite{Camelyon16}) and show that light microscopy images exhibit significantly higher redundancy, with the highest observed in real-world microscopic datasets.

Recent advancements in weakly-supervised multiple instance learning (MIL) have facilitated the use of WSIs for cancer diagnosis by requiring only slide-level labels, thereby reducing the need for exhaustive annotations \cite{gadermayr2024multiple}. Within this paradigm, graph-based MIL (GNN-MIL) models \cite{chan2023,bontempo2023} have shown promise by leveraging spatial relationships among patches to capture contextual information, representing patches as nodes and their interactions as edges. However, these methods are inherently designed for WSIs and cannot be directly applied to light microscopy images, where spatial coordinates are unknown and image redundancy is prevalent. Overcoming these limitations requires a specific approach that accommodates the unique characteristics of light microscopy images while preserving the benefits of graph-based contextual modeling.

To address the challenges of absent spatial coordinates and high redundancy, we propose \textbf{MicroMIL}, the first weakly-supervised MIL framework specifically designed for conventional light microscopy images. MicroMIL introduces a representative image extractor (RIE) that leverages deep cluster embedding (DCE) \cite{huang2014deep} to dynamically group redundant images and hard Gumbel-Softmax \cite{jang2016categorical} to select a representative image per cluster. These selected images serve as graph nodes, with edges formed using cosine similarity to capture contextual information among instances. While prior approaches have relied on statistical heuristics or ensemble-based methods \cite{Nguyen2021,Gandomkar2018}, the most related method \cite{kim2024leveraging} does not address critical challenges specific to light microscopy images. MicroMIL is explicitly designed to overcome these limitations through a graph-based formulation that operates without relying on spatial metadata.

By enabling end-to-end representative feature selection, MicroMIL jointly optimizes clustering and instance selection within a unified framework, ensuring that the most informative representations contribute to the final prediction. To achieve this, we propose an online redundancy-aware learning strategy that dynamically refines instance selection while maintaining feature diversity. The graph-based representation further enhances structural preservation by connecting similar nodes, mitigating the loss of spatial information. Extensive experiments on a real-world colon cancer dataset and the BreakHis dataset validate MicroMIL’s effectiveness, demonstrating significant gains in diagnostic accuracy and redundancy robustness compared to state-of-the-art MIL methods.

\begin{figure*}[!htb]
\centering
\includegraphics[width=0.8\linewidth, trim=0 150 500 0, clip]{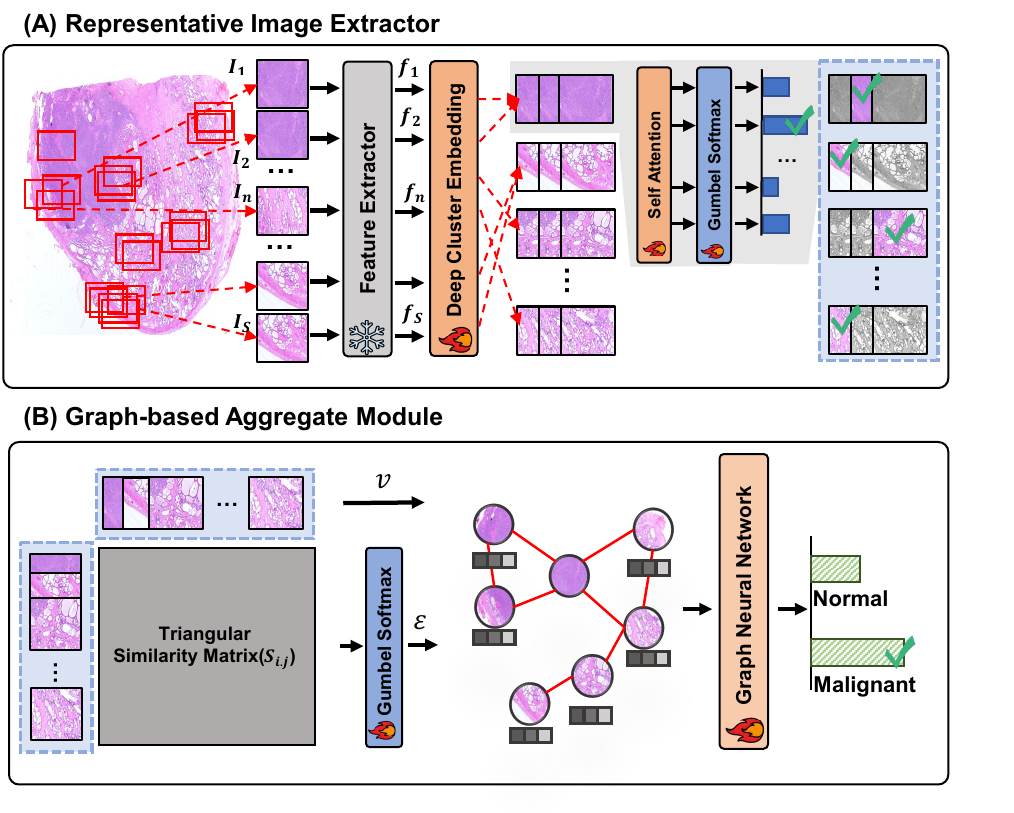}
\caption{Proposed end-to-end MicroMIL framework.}
\label{figure2}
\end{figure*}

\section{Methodology}

Each patient is treated as a \textit{bag}, and the corresponding light-microscope images as \textit{instances}, following the MIL paradigm. The goal is to predict patient diagnosis in a weakly-supervised setting without instance-level labels. We propose MicroMIL, a MIL framework for microscopic image analysis (see Figure \ref{figure2}). The framework consists of three components: (1) a frozen pre-trained feature extractor for generating image features, (2) a RIE that reduces redundancy by clustering similar images using DCE and selecting representatives via hard Gumbel-Softmax, and (3) a graph-based aggregation module, where nodes represent the selected represented images and edges are constructed using an upper triangular cosine similarity matrix to link similar nodes. Finally, a GNN captures contextual information for analysis.

\subsection{Representative Image Extractor}

In the embedding-based MIL framework, a frozen pre-trained feature extractor \( E \) maps each microscopic image \( I_s \) to a \( d \)-dimensional feature vector \( f_s = E(I_s) \), forming the feature set \( F = \{f_1, f_2, \ldots, f_S\} \), where \( S \) varies across patients. To group redundant images, we employ deep cluster embedding (DCE) \cite{Deep_Cluster_Embedding}, which iteratively assigns data points to clusters and refines cluster centers to minimize intra-cluster distances. Let \( \mu_c \in \mathbb{R}^d \) be the centroid of the \( c \)-th cluster, where \( c = 1, \ldots, C \). The soft assignment probability \( z_{s,c} \), indicating the likelihood of \( f_s \) belonging to the \( c \)-th cluster, is defined as:

\begin{equation}
z_{s,c} = \frac{\left( 1 + \| f_s - \mu_c \|^2 \right)^{-1}}{\sum_{j=1}^C \left( 1 + \| f_s - \mu_j \|^2 \right)^{-1}}, \quad Z \in \mathbb{R}^{S \times C}
\end{equation}

The DCE algorithm alternates between updating the centroids \( \mu_c \) and refining cluster assignments \( Z \) until convergence.

To select the most representative feature from each cluster, we use the hard Gumbel-Softmax function \cite{jang2016categorical}, which allows for end-to-end differentiability. Given logits \( X_x \) and Gumbel noise \( g_x \sim \text{Gumbel}(0,1) \), the hard Gumbel-Softmax function is defined as:

\begin{equation}
\text{HardGumbel}(X) = \text{one\_hot}\left(\arg\max_x (X_x + g_x)\right)
\end{equation}

Applying this function to feature-cluster interactions, the hard cluster assignments \( \tilde{Z} \) are determined as:

\begin{equation}
\tilde{z}_{s,c} = \text{HardGumbel}(s_{s,c}), \quad \tilde{Z} \in \mathbb{R}^{S \times C}
\end{equation}

where \( s_{s,c} = w^\top (f_s \odot z_{:,c}) \), with \( w \in \mathbb{R}^d \) being a learnable weight vector and \( \odot \) denoting element-wise multiplication.

The representative feature \( q_c \) of cluster \( c \) is computed as follows:
\begin{equation}
q_c = \sum_{s=1}^S \tilde{z}_{s,c} f_s, \quad Q \in \mathbb{R}^{C \times d}
\end{equation}
This process combines DCE for clustering and the hard Gumbel-Softmax for selecting representative features, ensuring inter-cluster separation and intra-cluster compactness. By focusing on representative features, this approach improves subsequent classification performance while reducing redundancy.

\subsection{Graph-based Aggregate Module}

To model relationships among clusters, we construct a graph \( G \), where nodes represent representative feature clusters and edges capture pairwise similarities. Given representative cluster embeddings \( Q = \{q_1, q_2, \ldots, q_C\} \), where \( q_c \in \mathbb{R}^d \), the pairwise similarity is computed using cosine similarity as \( S_{ij} = \frac{q_i^\top q_j}{\| q_i \| \| q_j \|} \), with \( \| q_i \| \) denoting the Euclidean norm of \( q_i \). A value of \( S_{ij} \) closer to 1 indicates higher similarity between clusters.

To retain only the most important relationships, we apply the same hard Gumbel-Softmax function to the similarity matrix:

\begin{equation}
\tilde{m}_{i,j} = \text{HardGumbel}(S_{ij}), \quad \tilde{M} \in \mathbb{R}^{C \times C}
\end{equation}

The resulting graph \( G = (V, E) \) is defined by nodes \( V = \{1, 2, \ldots, C\} \) and edges \( E = \{(i, j) \mid \tilde{m}_{i,j} > 0\} \).

Once the graph is constructed, a GNN propagates and refines the cluster embeddings. The initial node features are \( H^{(0)} = R \) and through \( L \) GNN layers, node embeddings are updated by aggregating information from neighboring nodes. The entire process is represented as:

\begin{equation}
y = \sigma \left( W_{\text{class}} \cdot \text{mean} \left( \text{GNN}(G, R) \right) \right)
\end{equation}

where \( \text{GNN}(G, R) \) represents the \( L \)-layer operations on \( G \), 
\( W_{\text{class}} \) is the classification weight matrix, 
\( \text{mean}(\cdot) \) aggregates node embeddings, and 
\( \sigma \) is the activation function. The entire framework, including DCE, RIE, and GNN, is trained end-to-end using binary cross-entropy (BCE) loss.

\section{Experiments and Results}

\subsection{Datasets}

\textbf{BreakHis} BreakHis \cite{Breakhis} is a widely used benchmark dataset for microscopy image analysis and cancer research. It comprises 7,909 images from 81 patients, with an average of 96.4 images per patient. Among these, 2,480 images (from 24 patients) are labeled as normal, while 5,429 images (from 57 patients) are labeled as malignant. The dataset spans multiple magnifications (40×, 100×, 200×, and 400×) and is collected under controlled conditions, which may limit its applicability to real-world scenarios.

\textbf{Real-world (Seegene) Dataset}  
Our real-world dataset, collected from the Seegene Medical Foundation, consists of 135,100 images from 899 patients, averaging 150.3 images per patient. The dataset was approved by both the foundation’s IRB (SMF-IRB-2020-007) and the KAIST IRB (KAIST-IRB-23-214). It includes 52,339 images (from 493 patients) labeled as normal and 82,761 images (from 406 patients) labeled as malignant, primarily at 100× and 200× magnifications, aligning with common clinical practices.

\subsection{Implementation Details}

For a fair comparison, we use ResNet18 pre-trained on ImageNet as the feature extractor for all models, with a hidden dimension of 128 for consistency. MicroMIL is tuned with a dropout rate of 0.5, a learning rate of $1 \times 10^{-3}$, and the Adam optimizer. For the final results, we use 36 clusters for the real-world (Seegene) dataset and 16 for BreakHis, given its smaller image count per patient. We experiment with 16 ($4^2$), 25 ($5^2$), and 36 ($6^2$) clusters and report the best-performing configuration. Performance differences across these settings are minimal and MicroMIL consistently outperforms all baselines. While online clustering (DCE) requires a predefined cluster count, we aim to explore automatic cluster number selection in future work. All models, implemented in PyTorch, are trained with two graph layers on an NVIDIA GeForce RTX 3080 GPU.

\subsection{Baselines Comparison}

\begin{table*}[!htb]
\scriptsize

\centering
\caption{Performance metrics of baselines and MicroMIL on Real-world (Seegene) and BreakHis datasets. Best results are in \textbf{bold} and second-best results are \underline{underlined}.}
\begin{tabular}{|l|ccc|ccc|}
\hline
 & \multicolumn{3}{c|}{\textbf{Real-world (Seegene) }} & \multicolumn{3}{c|}{\textbf{BreakHis}} \\ \hline
\textbf{Model} & \textbf{ACC} & \textbf{AUC} & \textbf{F1} & \textbf{ACC} & \textbf{AUC} & \textbf{F1} \\ \hline
ABMIL [ICML`18] \cite{ABMIL} & 0.9444 & 0.9764 & 0.9433 & 0.8929 & 0.8947 & 0.8805 \\ 
MS-DA-MIL [CVPR`20] \cite{MSDAMIL} & 0.9556 & 0.9829 & 0.9514 & 0.8929 & 0.9591 & \underline{0.9268} \\ 
DSMIL [CVPR`21] \cite{DSMIL} & 0.9444 & 0.9829 & 0.9440 & 0.8214 & 0.8947 & 0.8155 \\ 
CLAM [Nat BioMed`21] \cite{CLAM} & 0.9556 & 0.9873 & 0.9552 & 0.9286 & 0.9298 & 0.9181 \\
TransMIL [NeurIPS`21] \cite{TransMIL} & \underline{0.9778} & 0.9873 & \underline{0.9776} & 0.8929 & \underline{0.9825} & \underline{0.9268} \\ 
DTFD-MIL [CVPR`22] \cite{DTFD} & 0.9611 & \underline{0.9901} & 0.9607 & \underline{0.9286} & 0.9766 & 0.9222 \\ 
IBMIL [CVPR`23] \cite{ibmil} & 0.9611 & 0.9894 & 0.9606 & \underline{0.9286} & 0.9532 & 0.9181 \\ 
ACMIL [ECCV`24] \cite{ACMIL} & 0.9611 & 0.9893 & 0.9606 & 0.8929 & 0.9474 & 0.8857 \\ \hline 
\rowcolor{gray!20}
\textbf{MicroMIL } & \textbf{0.9922} & \textbf{0.9994} & \textbf{0.9925} & \textbf{0.9643} & \textbf{0.9942} & \textbf{0.9730} \\ \hline
\end{tabular}

\label{table:metrics}
\end{table*}

\Cref{table:metrics} shows that MicroMIL consistently surpasses baseline MIL models across all evaluated metrics on both real-world and public datasets. This performance advantage stems from the MicroMIL's specific design to address two key challenges of light microscopy datasets: image redundancy and no absolute position. In contrast, existing MIL models are designed for WSIs obtained from scanners, thereby without the need to account for these characteristics. By effectively tackling the unique characteristics present, MicroMIL proves to be exceptionally well-suited for patient diagnosis using light microscopy images.

\subsection{Impact of Representative Image Extractor}

\begin{figure*}[!htb]
    \centering
    \begin{minipage}{0.49\linewidth}
        \centering
        
        \includegraphics[width=0.9\linewidth]{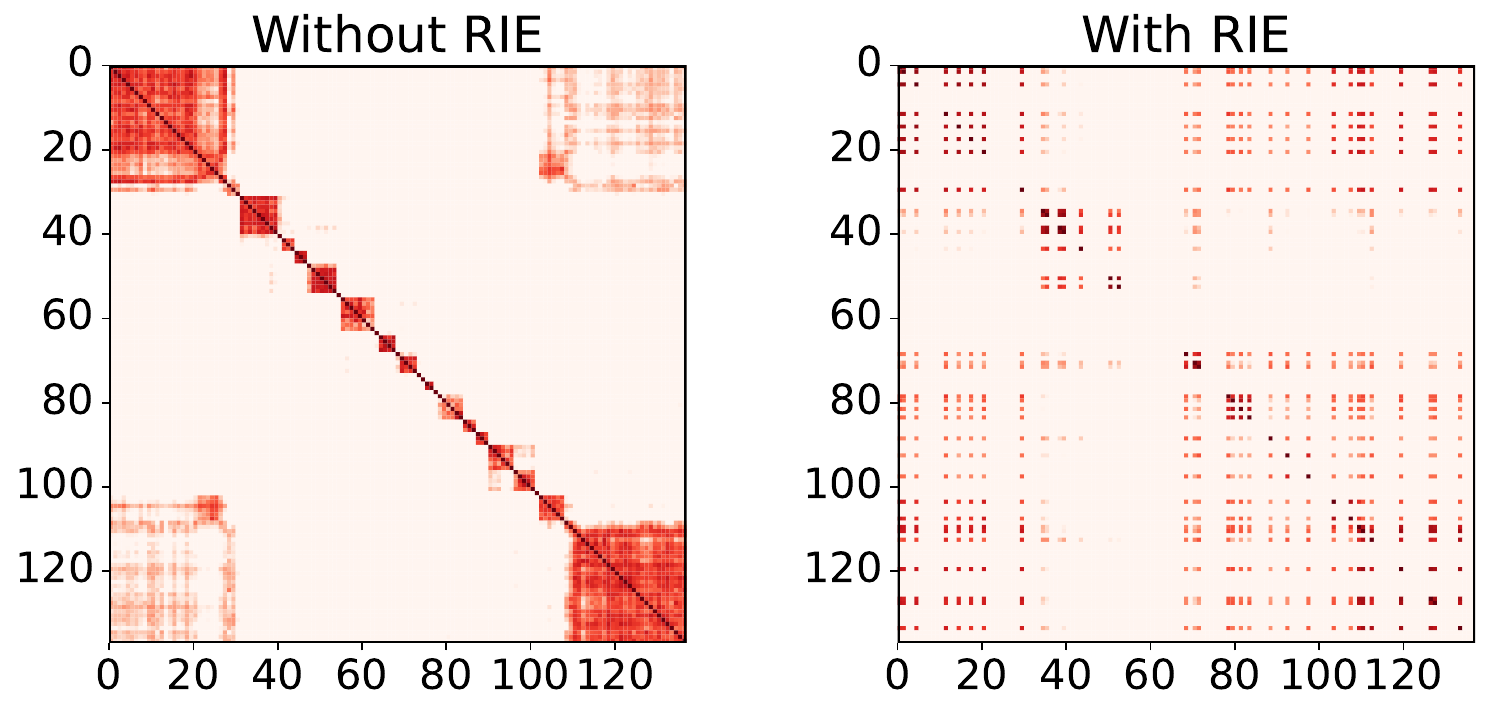}
        \caption{Similarity-based edge formation probabilities ($\geq 0.8$) heatmaps for real-world (Seegene) data, comparing cases \textit{without (Left)} and \textit{with (Right)} the RIE.}
        \label{heatmap}
    \end{minipage}
    \hfill
    \begin{minipage}{0.49\linewidth}
        \centering

        \includegraphics[width=1\linewidth]{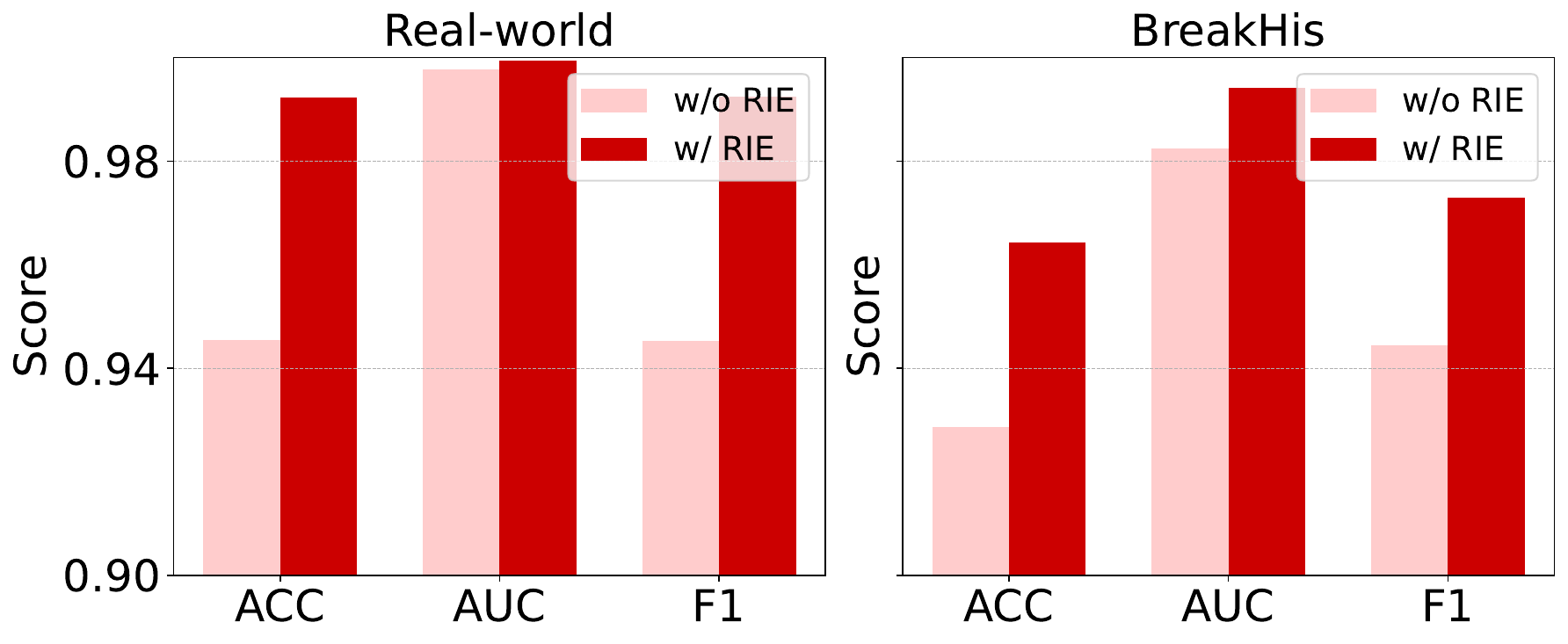}
        \caption{Performance metrics \textit{without (w/o)} and \textit{with (w/)} the RIE on Real-world (Seegene) and BreakHis datasets. }
        \label{cluster_table2}
    \end{minipage}
\end{figure*}

In histopathology, patient-level prediction requires information exchange across diverse tissue regions \cite{ahn2021multiregion,halperin2019leveraging}, but redundant images hinder this process. To address this, we propose the Representative Image Extractor (RIE), which selects representative images from visually similar clusters to enhance diversity and improve predictive performance. \Cref{heatmap} (\textit{without} RIE) shows that redundancy limits diverse interactions and leads to performance degradation (\Cref{cluster_table2}), which RIE successfully mitigates by filtering redundant instances. The effect varies depending on dataset redundancy, with smaller improvement in BreakHis due to lower redundancy, whereas the real-world dataset shows a larger improvement, demonstrating RIE’s effectiveness in handling highly redundant data.

\subsection{Connecting Relative Neighborhood Nodes Method}
\begin{figure*}[!htb]
    \centering
    \begin{minipage}{0.49\linewidth}
        \centering
        
        \includegraphics[width=1\linewidth]{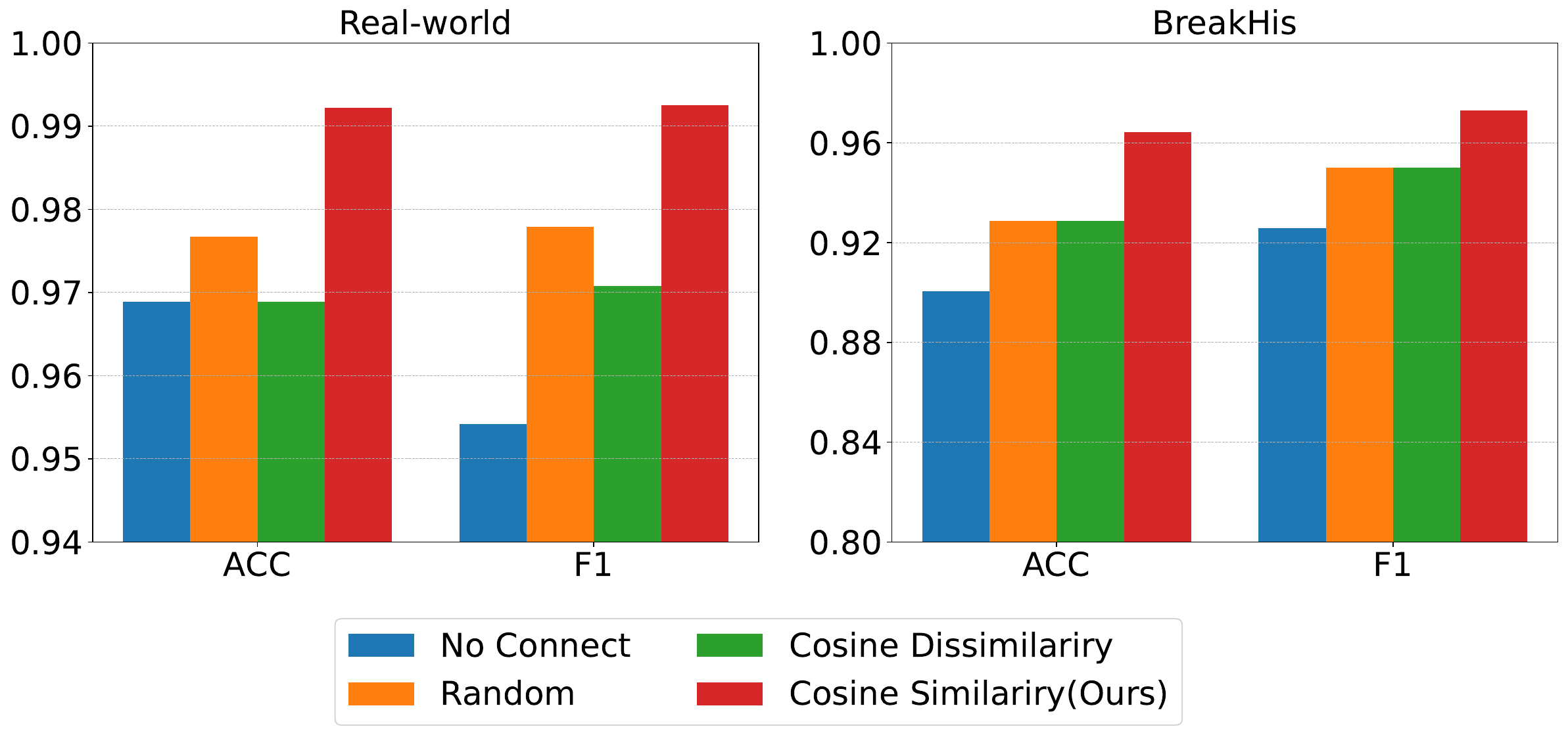}
        \caption{Performance metrics of different edge generation methods}
        \label{method2}
    \end{minipage}
    \hfill
    \begin{minipage}{0.49\linewidth}
        \centering

        \includegraphics[width=1\linewidth]{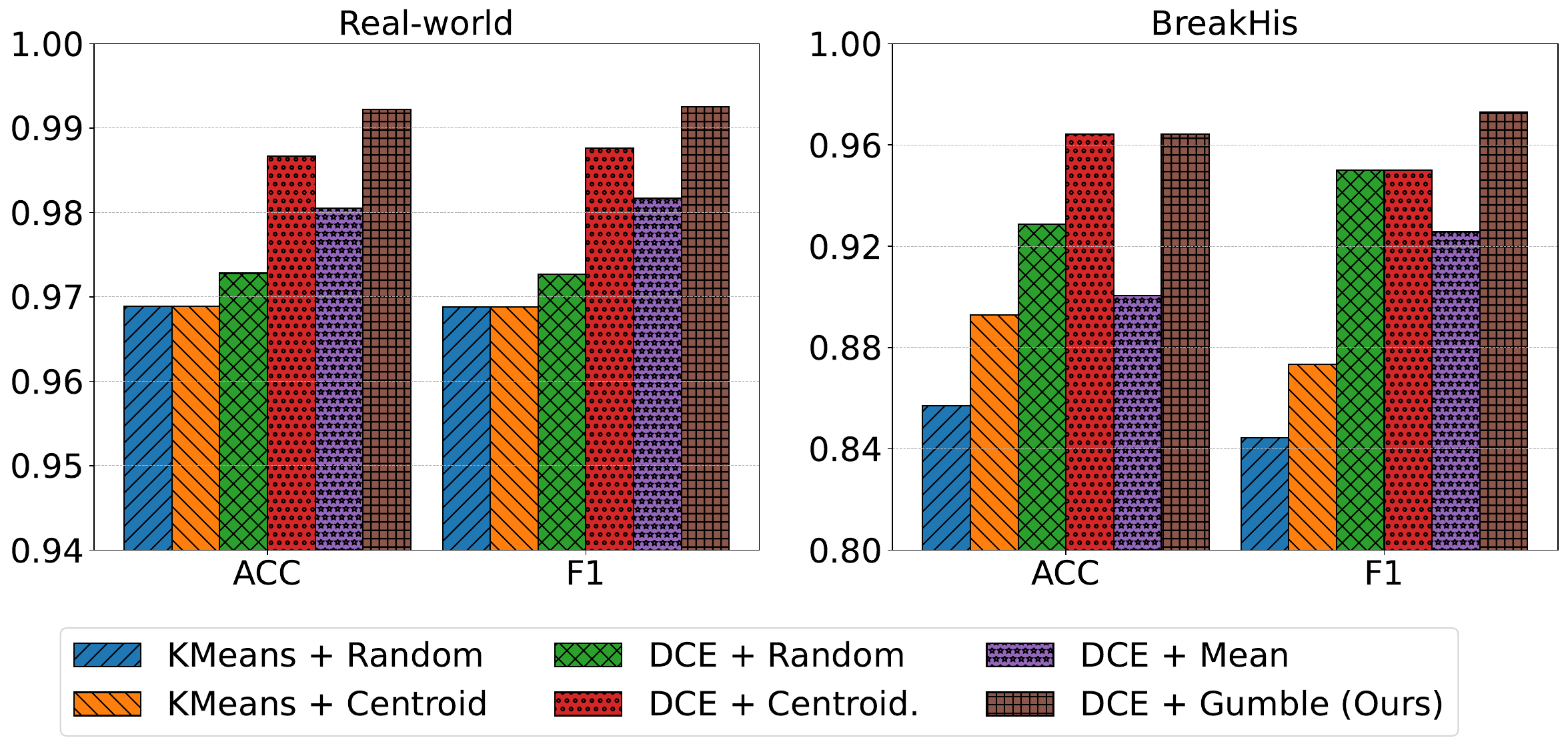}
        \caption{Performance metrics of different representative image extractor methods}
        \label{method1}
    \end{minipage}
\end{figure*}

Light microscopy images lack spatial information, necessitating effective edge construction. We utilize a GNN-based method that connects nodes to their most similar neighbors. As shown in Figure \ref{method2}, performance drops without connections due to a lack of contextual relationships. Random connections lead to weak interactions, while cosine similarity-based edges, linking highly similar nodes, capture meaningful relationships and outperform random or reverse-similarity (1/similarity) edges. This highlights the importance of leveraging similarity to enhance context integration and predictive performance.

\subsection{Ablation Studies}

\textbf{Representative Image Extractor Methods.} To enhance feature learning, we adopt an online approach using DCE and Gumbel-Softmax for selecting influential representative images. As shown in \Cref{method1}, a comparison of clustering methods (KMeans, DCE) and selection strategies (Random, Mean, Centroid, Gumbel-Softmax) reveals that offline KMeans underperforms, while online DCE + Gumbel-Softmax achieves superior results. Notably, Gumbel-Softmax outperforms Mean and Centroid by better highlighting key images, demonstrating the effectiveness of online clustering and selection in optimizing feature learning. This online approach enables dynamic adaptation, ensuring robust feature representations and reducing the risk of suboptimal cluster assignments.

\begin{table}[!hbt]
\centering
\scriptsize
\setlength{\tabcolsep}{3pt} 
\caption{Robustness to redundancy in baseline models and MicroMIL. Arrows indicate data flow direction (train $\rightarrow$ test). We count images exceeding the 0.995 redundancy threshold (\Cref{figure1}, middle) per patient, then select the top 10\% (\textbf{T10}) and bottom 10\% (\textbf{B10}) of patients. Best results are in \textbf{bold}.}
\begin{tabular}{|l|ccc|ccc|ccc|}
\hline
 & \multicolumn{3}{c|}{\textbf{(1) T10 $\rightarrow$ B10}} & \multicolumn{3}{c|}{\textbf{(2) B10 $\rightarrow$ T10}} & \multicolumn{3}{c|}{\textbf{(3) T10 $\rightarrow$ T10}} \\ \hline
\textbf{Model} & \textbf{ACC} & \textbf{AUC} & \textbf{F1} & \textbf{ACC} & \textbf{AUC} & \textbf{F1} & \textbf{ACC} & \textbf{AUC} & \textbf{F1} \\ \hline
ABMIL & 0.8090 & 0.8592 & 0.7966 & 0.9213 & 0.9592 & 0.9210 & 0.9091 & 0.9229 & 0.9089 \\ 
MS-DA-MIL & 0.9101 & 0.9526 & 0.9248 & 0.9213 & 0.9642 & 0.9248 & 0.9091 & 0.9348 & 0.9209 \\ 
DSMIL & 0.9101 & 0.9474 & 0.9092 & 0.9326 & 0.9755 & 0.9319 & 0.9091 & 0.9438 & 0.9089 \\ 
CLAM & 0.9326 & 0.9796 & 0.9315 & 0.9213 & 0.9770 & 0.9194 & 0.9318 & 0.9521 & 0.9318 \\ 
TransMIL & 0.9213 & 0.9776 & 0.9212 & 0.8989 & 0.8526 & 0.8963 & 0.9318 & 0.8854 & 0.9309 \\ 
DTFD-MIL & 0.9438 & 0.9658 & 0.9428 & 0.9213 & 0.9750 & 0.9203 & 0.9318 & 0.9583 & 0.9318 \\ 
IBMIL & 0.9326 & 0.9709 & 0.9319 & 0.9326 & 0.9719 & 0.9319 & 0.9318 & 0.9458 & 0.9315 \\ 
ACMIL & 0.9434 & 0.9704 & 0.9431 & 0.9213 & 0.9689 & 0.9210 & 0.9318 & 0.9521 & 0.9315 \\ \hline
\rowcolor{gray!20}
\textbf{MicroMIL} & \textbf{0.9663} & \textbf{0.9842} & \textbf{0.9630} & \textbf{0.9551} & \textbf{0.9801} & \textbf{0.9542} & \textbf{0.9545} & \textbf{0.9958} & \textbf{0.9524} \\ \hline
\end{tabular}
\label{table:image_redundancy_shift}
\end{table}

\textbf{Robustness on Image Redundancy Shift.} To evaluate MicroMIL’s robustness to image redundancy, we set a similarity threshold of 0.995 to identify redundant image pairs. \Cref{table:image_redundancy_shift} shows that baseline MIL methods degrade under extreme redundancy settings (B10$\rightarrow$T10, T10$\rightarrow$T10), while MicroMIL consistently maintains high performance. Even in low-redundancy simulations (T10$\rightarrow$B10), MicroMIL outperforms all baselines. These results confirm MicroMIL’s ability to extract critical features and remain effective in any scenario.

\section{Conclusion}

We introduce MicroMIL, the first weakly-supervised MIL framework specifically designed for conventional light microscopy images, addressing the limitations of GNN-MIL approaches that rely on spatial coordinates and exhibit low redundancy tolerance. By integrating deep cluster embedding (DCE) and hard Gumbel-Softmax, MicroMIL effectively reduces redundancy and selects representative instances, enabling a graph-based representation without requiring absolute spatial positioning while explicitly modeling contextual cues. Experiments on the real-world and BreakHis datasets demonstrate state-of-the-art performance, improving diagnostic accuracy while maintaining robustness to redundancy. MicroMIL offers a scalable, spatially agnostic solution that advances weakly-supervised MIL for microscopy imaging in resource-constrained settings.

\begin{credits}
\subsubsection{\ackname} This work was supported by the National Research Foundation of Korea (NRF) grant funded by the Korea government (MSIT) (No.RS-2022-NR068758).

\subsubsection{\discintname} The authors have no competing interests to declare that are
relevant to the content of this article.
\end{credits}

%
%
%
\bibliographystyle{splncs04}
\bibliography{LatexSource-0707/Paper-0707}

\end{document}